\documentclass{article}
\usepackage{spconf,amsmath,graphicx}

\usepackage{times}
\usepackage{epsfig}
\usepackage{amssymb}
\usepackage{booktabs}

\usepackage{amsmath,amsfonts,bm}

\def\eqref#1{equation~\ref{#1}}

\def\1{\bm{1}}

\def\vmu{{\bm{\mu}}}

\def\vp{{\bm{p}}}

\def\vv{{\bm{v}}}

\DeclareMathAlphabet{\mathsfit}{\encodingdefault}{\sfdefault}{m}{sl}
\SetMathAlphabet{\mathsfit}{bold}{\encodingdefault}{\sfdefault}{bx}{n}

\def\sA{{\mathbb{A}}}

\newcommand{\R}{\mathbb{R}}

\usepackage[acronym]{glossaries}
\usepackage{fancyhdr}
\usepackage{multicol}
\usepackage{nicefrac}
\usepackage{transparent}
\usepackage{watermark}
\graphicspath{ {./img/} }

\usepackage{additional_stuff}

\renewcommand{\paragraph}[1]{\noindent\textbf{#1}\quad}
\title{Biologically-Inspired Continual Learning of Human Motion Sequences}
\name{Joachim Ott, Shih-Chii Liu\thanks{
 Part of the work was done at Starmind AG by the first author. The subsequent work at University of Zurich was partially supported by the Swiss National Science Foundation CA-DNNEdge (208227).}}
\address{Institute of Neuroinformatics, University of Zurich and ETH Zurich, Switzerland \\}

    \newacronym{dnn}{DNN}{Deep Neural Network}
    \newacronym{rnn}{RNN}{Recurrent Neural Network}
   \newacronym{cnn}{CNN}{Convolutional Neural Network}
    \newacronym{lstm}{LSTM}{Long Short-Term Memory}
     \newacronym{gmm}{GMM}{Gaussian Mixture Model}
     \newacronym{vae}{VAE}{Variational Auto-Encoder}
      \newacronym{fid}{FID}{Fŕechet Inception Distance}

\begin{document}
\ninept  
\maketitle
\thispageheading{\vspace{-10mm}   \centering\transparent{0.3}This paper has been accepted to the\\ IEEE International Conference on Acoustics, Speech, and Signal Processing (ICASSP), Rhodes Island, Greece, 2023}

\thiswatermark{

\put(0,-690){\parbox{\textwidth}{%
        \transparent{0.5}\textcopyright 2023 IEEE. Personal use of this material is permitted. Permission from IEEE must be obtained for all other uses, in any current or future media, including reprinting/republishing this material for advertising or promotional purposes, creating new collective works, for resale or redistribution to servers or lists, or reuse of any copyrighted component of this work in other works.}
    }%
    }

\begin{abstract}
This work proposes a model for continual learning on tasks involving temporal sequences, specifically, human motions. It improves on a recently proposed brain-inspired replay model (BI-R) by building a biologically-inspired conditional temporal variational autoencoder (BI-CTVAE), which instantiates a latent mixture-of-Gaussians for class representation. 
We investigate a novel continual-learning-to-generate (CL2Gen) scenario where the model generates motion sequences of different classes. The generative accuracy of the model is tested over a set of tasks.
The final classification accuracy of BI-CTVAE on a human motion dataset after sequentially learning all action classes is 78\%, which is 63\% higher than using no-replay, and only 5.4\% lower than a state-of-the-art offline trained GRU model. 
\end{abstract}
%

\section{Introduction}

Humans and other animals show an amazing ability to learn continuously without forgetting, \ie, they do not lose their ability to perform old skills even after learning new skills for new tasks~\cite{bremner2012multisensory,tani2016exploring,parisi2019continual}.
This ability implies that the old knowledge is retained and can be used in new situations~\cite{parisi2019continual}.
In the case of procedural knowledge like motor movements and motions, \eg, in training for a sport like tennis, humans learn and refine their strokes not only through practice, but also through mental rehearsal and observation \cite{mizuguchi2017changes,nakano2017understanding,neuman2013direct,ste2012observation}. To consolidate the acquired procedural knowledge and additional precision improvements, motions are re-generated or replayed during sleep, as observed in rodents and humans \cite{ramanathan2015sleep,cousins2016cued}. This knowledge can then be consolidated into long-term memory in the hippocampus~\cite{cousins2016cued}.

Common machine learning models are not able to retain or transfer knowledge as biological brains can like described above, 
since they suffer from catastrophic forgetting when trained on new classes \cite{mccloskey1989catastrophic,french1999catastrophic}. Continual learning explores solutions to overcome these limitations. 

Various methods for continual learning have been proposed~\cite{parisi2019continual}. The replay continual learning method, in particular  a biologically plausible version of generative replay (BI-R), is one of the methods that perform well in task-, class-, and domain-incremental learning \cite{van2020brain}, hence it is used as a basis of our work on continual learning to generate motion. One of its advantages is that generative replay removes the necessity of storing any samples or exemplars of previously learned tasks. 
Motion is a sequential data type consisting of individual poses, thus a sequential model is required. So far, only a few continual learning studies have tested or developed models for tasks involving sequential or time-series data like text or video, \eg, \cite{cossu2021continual,ehret2020continual,sodhani2020toward,ororbia2020continual,sun2019lamol,duncker2020organizing,kao2021natural}. 

In most studies that use a generative replay approach, the generated samples are just a tool to maintain classification accuracy. These generated samples can even be of low quality but are still useful for producing good incremental learning results on classification tasks \cite{van2020brain}. In contrast, good generative performance, \eg, in imitating or reproducing motions, is very important in many real life situations.
This biology rooted feature of continual learning to generate is still under-explored in machine learning. 
Therefore, this work addresses a continual learning setting that is closer related to the situation that biological brains encounter, which we call {\it continual-learning-to-generate} (CL2Gen).

To address CL2Gen on motion, we propose a new model called BI-CTVAE that uses a biologically-inspired conditional temporal variational autoencoder. 
We chose generative replay for retaining past knowledge and propose a powerful representation and generative architecture that allows generation of entire sequences of a motion class 
from a single latent space vector. The model is benchmarked on HumanAct12, a human motion dataset with 12 tasks of different action classes.

The key contributions of this work are as follows:
\begin{itemize}
    \item A new continual learning study which focuses on human motion. To the best of our knowledge, this work is one of the first to propose continual learning to generate human motion sequences.
    \item A novel model (BI-CTVAE) for biologically-inspired CL2Gen using mode-based generative replay and applied on tasks involving temporal sequences.
    \item Adapting HumanAct12, a state-of-the-art human motion dataset, used here for training the CL2Gen models. 
    The adapted dataset can be used for future incremental learning work concerning human motion generative performance.
\end{itemize}

\section{CL2Gen of Motions with Conditional Temporal VAE} 

\begin{figure*}[h]
\begin{center}

\includegraphics[width=\textwidth]{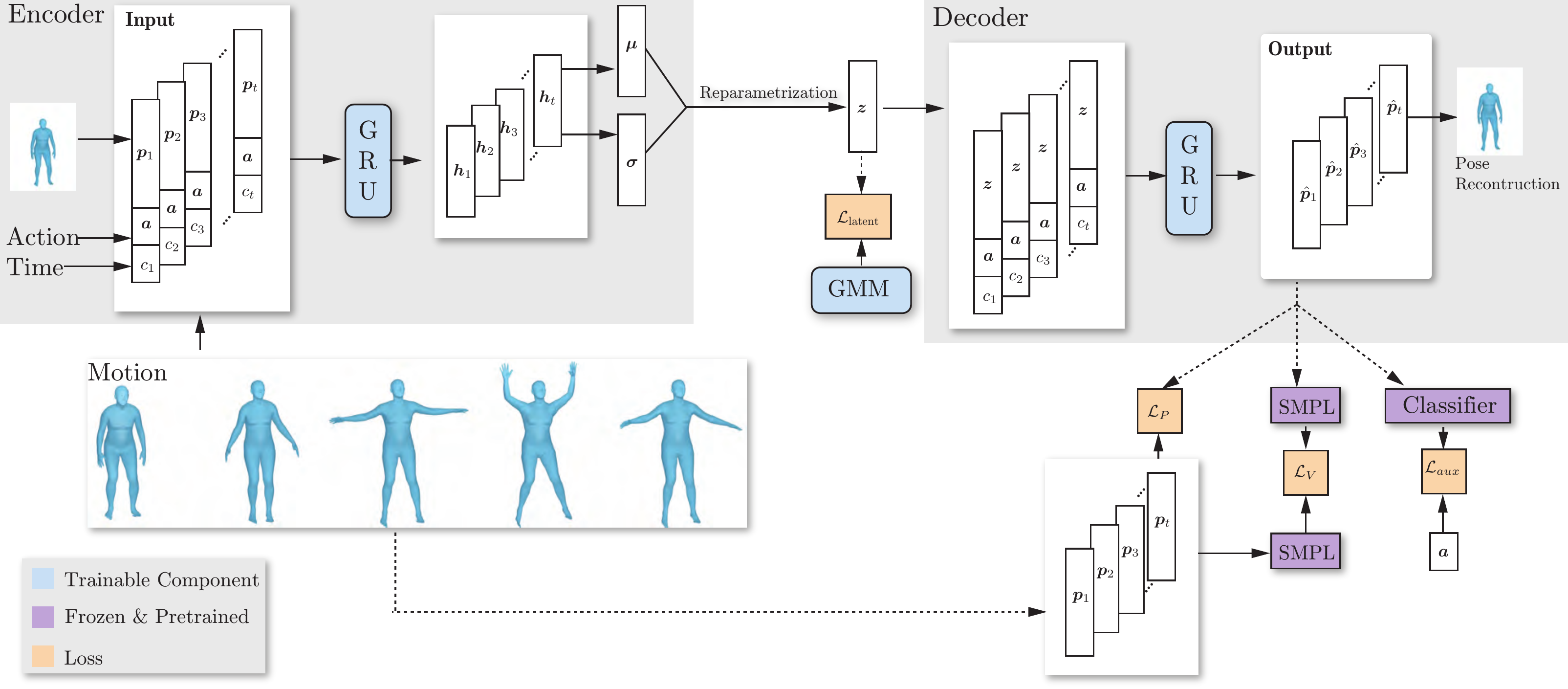}
\end{center}
\vspace*{-6mm}
\caption{Overview of BI-CTVAE components. \textbf{Encoder}: Input is a motion sequence of poses $\bm{p}_1$...$\bm{p}_T$, each  concatenated with the action label $a$ and a time index $c_1$...$c_T$. The input is first processed by a multi-layered GRU network. The hidden state of the last layer is used to calculate the latent vectors,  $\vmu$ and $\bm{\sigma}$. These are used to sample a motion latent representation $\bm{z}$. \textbf{Decoder}: Input is a sequence of repeated $\bm{z}$, each concatenated with action label $a$ and time index $c_1$...$c_T$. Decoder output is the reconstructed motion. \textbf{GMM}: The per-class \gls{gmm} components allow sampling of classes learned in previous tasks.  }
\vspace*{-5mm}
\label{fig:main_model}
\end{figure*}

Here, we describe the basics of human motion representation, CL2Gen, and our BI-CTVAE model. 

\subsection{Prerequisites}
\paragraph{Class-incremental continual learning}
In the class-incremental scenario of continual learning, a model is trained on a sequence of tasks $k_1$...$k_K$ from a task sequence $K$. Each task involves training on samples from one or multiple $n$ classes $a_1$...$a_n$. The model is not aware of the task boundaries, hence, \eg,  an output prediction, in a prediction task, contains the probabilities for all classes seen so far. In CL2Gen, no task info is provided to the model either, hence it is similar in this regard to class-incremental continual learning.

\paragraph{Motion action definition}
A motion $\bm{M}$ of action $a$ from the set of actions $\sA$ is a sequence of body pose frames $\bm{p}_1$...$\bm{p}_T$ with total motion duration $T$. 
A pose frame, $\bm{p}_t$, contains information of the position / displacement $\bm{d}$ of a low-dimensional skeleton pose $\bm{r}$ in space at time point $t$. The skeleton is made of a number of joints $J$, where each joint is represented with dimensionality $B$, hence a pose frame can be defined as $\bm{r} \in \R^{J \times B}$.

\paragraph{Motion representation in latent space}
An action represented by a motion $\bm{M}$ may be encoded into a latent variable $\Pr(\bm{z} \mid \bm{M})$, by first encoding the entire sequence into $\vmu^{(\bm{M})}$ and $\bm{\sigma^{(M)}}$. A new sample, $\bm{z}$, can be obtained by sampling from 
$\mathcal{N}(\bm{\mu^{(M)}}, \bm{\sigma^{(M)}})$ (see Fig.~\ref{fig:main_model}).

\paragraph{CL2Gen}
In the CL2Gen scenario, the model is trained incrementally on tasks where each task is represented by a set of action classes. The goal is for the model to maintain the ability to generate representative samples of classes of previous tasks even after training on the new classes within each new task. To evaluate the samples, either a classifier trained on real samples and / or other metrics are used, as described in Section \ref{sec_exp_metrics}.
In contrast to most previous work on continual learning, this work focuses not only on classification but also on the accuracy of \textit{the model's generative performance}. We emphasize this, because it was previously shown that maintaining classification performance is possible even with low-quality generated replay samples \cite{van2020brain}. The next section describes the BI-CTVAE model and how it maintains its generative performance accuracy.

\subsection{BI-CTVAE}
To successfully learn to generate motions based on an action label and to maintain the generative performance over several tasks, a generative model with modifications enabling continual learning is required.
State-of-the-art conditional motion generation uses some combination of motion diffusion models~\cite{tevet2022human}, implicit neural representation \cite{cervantes2022implicit}, and transformers \cite{petrovich21actor}. These models are non-causal, \ie, they do not encode time and rely on some form of positional encoding of the input, which adds another level of non-biological plausibility. For this reason, in this work, we chose to use an \gls{rnn} as the basic architecture similar to \cite{chuan2020action2motion}. 
Since the purpose of our model is to not only encode but also to generate sequences of specific actions, GRUs\cite{cho2014properties} are used for both the encoder and the decoder of a conditional temporal VAE. 
Similar to \cite{cervantes2022implicit}, we adopt the framework and data representation of \cite{petrovich21actor}, however,  GRU networks are used instead of transformers as the base of our BI-CTVAE model. 
We extend the base model to be suitable for incremental learning of human motions using generative replay. 

Figure \ref{fig:main_model} shows the entire pipeline. The input to the model consists of the frame-wise motion encoding  concatenated with the frame index and action label. The decoder GRU can use a latent vector to generate a motion sequence. 
During training we use a set of class-specific trainable parameters as components of a Gaussian Mixture Model (GMM): for every class $a$ we have a trainable $\vmu^a$ and $\bm{\sigma}^a$. $\mathcal{X}$ is the collection of all trainable parameters of their means and standard deviations.

\paragraph{Motion representation}
The full pose frame $\bm{p}$ consists of the body pose $\bm{r}$ with 23 joints + global rotation and the displacement $\bm{d} \in \R^{3}$ of the root joint.  As previously established in \cite{petrovich21actor,cervantes2022implicit}, we transform the 3D representation of each of the 23 joints by adopting the method used in \cite{zhou2019continuity}, \ie, the 23 joint rotations and one global rotation are transformed into a 6D matrix representation, resulting in $\bm{r} \in \R^{24 \times 6}$.

\subsection{Mixture generative replay}
\label{sec:mix_replay}
For incremental learning with generative replay, the model most recently trained on task $k_n$ is used to generate replay samples from all previous classes (and tasks) that the model has been trained on, \ie $k_1$...$k_{n}$. 
These generated replay samples are combined with real samples of the next task $k_{n+1}$, and the model is trained on this mixed dataset. For task $k_{n+2}$, a completely new set of replay samples is generated from the newly trained model on $k_{n+1}$.

\subsection{Training}
\label{sec:training_loss}
The training objective is chosen such that the VAE learns via an encoder component to transform an input sequence $\bm{M}$ into a single vector of stochastic latent variable $\bm{z}$ and, in-turn, map $\bm{z}$ via a decoder component to a reconstruction $\hat{p}_1$...$\hat{p}_t$ of the input sequence. 

The total loss term used in training the CL2Gen model, $\mathcal{L}$, consists of a combination of different loss functions.  
In addition to the usual loss functions used in conditional motion generation and VAE training, we introduce additional loss functions for the CL2Gen scenario and add an auxiliary loss function based on the classifier output. We also adapt common VAE loss terms so that they are suitable for incremental learning. These loss functions are described next.

\paragraph{Latent loss}
The standard regularization term for training VAEs is the Kullback Leibler Divergence (KL) loss, which forces the latent space to be closer to a Gaussian distribution with unitary variance $\displaystyle \mathcal{N} ( \vmu , \bm{I})$, where $\mu$ = 0 in all dimensions and $\bm{\sigma}$ is the identity matrix. Since we use a separate mode for every class, we  modify the regularization term as follows:
\begin{align}
    & \mathcal{L}_{\text{latent}} (\bm{M},a,\bm{\phi}, \mathcal{X}) =
    \notag \\
    & \frac{1}{2} \sum_{j=1}^{256} \left( 1 + \log(\sigma_{j}^{(\bm{M})^2})-\log (\sigma_{j}^{a^2}) - \right.     \notag
    \left. \frac{\left( \mu_{j}^{(\bm{M})} -\mu_{j}^{a} \right)^2 + \sigma_{j}^{(\bm{M})^2} }{\sigma_{j}^{a^2}} \right)
    \end{align}
where $\bm{M}$ is the input motion sequence, $a$ the class label, $\mathcal{X}$ is the set of trainable means and  standard deviations, $\bm{\mu^{(M)}}$ and $\bm{\sigma^{(M)}}$ are calculated from $\bm{M}$ by the model with parameters $\bm{\phi}$, $\vmu^a$ and $\bm{\sigma}^{a}$ are the trainable mean and standard deviation of the mode corresponding to class $a$, $\mu_{j}^{a}$ and $\sigma_{j}^{a}$
 are $j$th elements respectively of $\vmu^{a}$ and $\bm{\sigma^{a}}$.
For additional details, we refer readers to Eq. 10 in \cite{van2020brain},  which forms the basis to this loss function.
\\
\textbf{Vertex loss} uses a pretrained and frozen SMPL model \cite{loper2015smpl} to extract the root-centered vertices of the mesh $\vv$ and $\hat{\vv}$ from poses $\vp$ and $\hat{\vp}$: 
${\vv}_t = \text{SMPL}({\vp}_t); 
\hat{\vv}_t = \text{SMPL}(\hat{\vp}_t)$.
We then calculate an L2 loss for every pose vertex in the sequence:
     $\mathcal{L}_{\text{V}} = \sum_{t=1}^{T} || \vv_t - \hat{\vv}_t ||_{2}^{2}$.
\\
\textbf{Reconstruction loss} uses the L2 loss on the reconstructed pose $\hat{\bm{p_t}}$ as follows:
  $\mathcal{L}_{\text{P}} = \sum_{t=1}^{T} || \vp_t - \hat{\vp}_t ||_{2}^{2}$.

\paragraph{Auxiliary loss} $\mathcal{L}_{\text{aux}}$ is the cross entropy loss calculated using the ground truth action label $a$ and the predicted label $\hat{a}$ obtained from a pretrained and frozen classifier on the reconstructed motion $\hat{\bm{M}}$.
\textbf{Final loss} consists of the loss components above and scaling factors: we include a hyperparameter $\lambda_{\text{latent}}$ for $\mathcal{L}_{\text{latent}}$, similar to the $\lambda_{KL}$ term in \cite{petrovich21actor} and $\beta$ in \cite{higgins2016beta}. 
We also include a hyperparameter $\lambda_{\text{aux}}$ for $\mathcal{L}_{\text{aux}}$.
The final loss is
\begin{equation}
  \mathcal{L} = \mathcal{L}_{\text{V}} + \mathcal{L}_{\text{P}} + \lambda_{\text{latent}} \mathcal{L}_{\text{latent}} + \lambda_{\text{aux}}
    \mathcal{L}_{\text{aux}}
\end{equation}
which is optimized with respect to $\phi$, $\mu^{a}$, and $\sigma^{a}$

\section{Experiments}
\begin{table*}[h]
\caption{Results for offline and CL2Gen (6 tasks) trained models (BI-CTVAE and BI-CTVAE aux) on HumanAct12. The  $\rightarrow$ symbol indicates metrics that should align with the values of the ground truth. $\pm$ is the $95\%$ confidence interval. Models trained in a CL2Gen setting with 
auxiliary loss and more replay samples perform better than those trained without replay. * indicates replay samples of previous tasks.}
\label{humanact12_results}
\begin{center}
\begin{tabular}{lllllll}
\multicolumn{1}{c}{\bf Model} 
&\multicolumn{1}{c}{\bf Replay} &\multicolumn{1}{c}{\bf Samples*} &\multicolumn{1}{c}{\bf Accuracy$\uparrow$}&\multicolumn{1}{c}{\bf FID$\downarrow$}    &\multicolumn{1}{c}{\bf Diversity$\rightarrow$}  &\multicolumn{1}{c}{\bf Multimodality$\rightarrow$}
\\ \hline 
{\bf Ground Truth} {\tiny \cite{chuan2020action2motion}} &  No & - & $99.7^{\pm0.01}$& $0.092^{\pm0.007}$  & $6.85^{\pm0.05}$ & $2.45^{\pm0.04}$\\
 \hline 

GRU (offline) & No & - & $74.2^{\pm1.04}$& $0.71^{\pm0.00}$  & $6.63^{\pm0.02}$ & $4.31^{\pm0.04}$\\
Ours (offline)  & No & -& $84.0^{\pm1.37}$ & $0.89^{\pm0.01}$  & $6.69^{\pm0.02}$ & $3.48^{\pm0.02}$\\
\hline 

GRU & No & - & $13.1^{\pm1.38}$& $11.18^{\pm0.24}$  & $5.91^{\pm0.05}$ & $5.47^{\pm0.05}$\\
BI-CTVAE & No & - & $14.8^{\pm1.00}$& $13.18^{\pm0.24}$  & $5.22^{\pm0.02}$ & $4.50^{\pm0.04}$\\
  &$\nicefrac{1}{16}$ & Real & $42.2^{\pm2.43}$& $\bm{1.73^{\pm0.03}}$  & $\bm{6.60^{\pm0.03}}$ & $4.72^{\pm0.03}$\\[0.1cm]
  &$\nicefrac{1}{16}$ & Gen & $43.1^{\pm4.62}$& $2.31^{\pm0.02}$  & $6.55^{\pm0.03}$ & $4.48^{\pm0.02}$\\[0.1cm]
&$\nicefrac{1}{5}$ & Gen & $48.9^{\pm2.49}$& $2.21^{\pm0.02}$  & $6.54^{\pm0.02}$ & $4.27^{\pm0.02}$\\[0.1cm]

\hline 
GRU aux  & $\nicefrac{1}{16}$ & Gen & $21.9^{\pm1.58}$ & $7.25^{\pm0.16}$  & $6.06^{\pm0.02}$ & $5.34^{\pm0.03}$\\
 & $\nicefrac{1}{5}$ & Gen & $42.5^{\pm2.83}$ & $5.52^{\pm0.13}$  & $6.06^{\pm0.03}$ & $5.12^{\pm0.03}$\\
BI-CTVAE aux& $\nicefrac{1}{16}$ & Real & $37.5^{\pm2.74}$& $3.33^{\pm0.06}$  & $6.75^{\pm0.02}$ & $5.29^{\pm0.07}$\\
 &$\nicefrac{1}{16}$  &  Gen & $61.6^{\pm2.37}$& $3.67^{\pm0.04}$  & $6.56^{\pm0.02}$ & $4.15^{\pm0.01}$\\
 &$\nicefrac{1}{5}$ 
& Gen & $\bm{78.6^{\pm1.40}}$& $2.70^{\pm0.01}$  & $6.57^{\pm0.03}$ & $\bm{3.07^{\pm0.03}}$\\
\end{tabular}
\end{center}
\vspace{-6mm}
\end{table*}

\subsection{Dataset}\label{sec_exp_dataset}

 HumanAct12 is a human motion dataset~\cite{chuan2020action2motion} derived from the PHSPD dataset~\cite{zou20203d,yang2019task}. It consists of 1191 motion sequences from 12 different action categories.  
Note: This dataset does not provide  separate training and test sets. For this work, we have created the necessary training set for our incremental learning experiments by splitting the 12 categories into individual tasks. In all our experiments, except offline learning, we use 2 classes per task, hence a total of 6 tasks.

\subsection{Hyperparameters and evaluation metrics}\label{sec_exp_metrics}
In all experiments  $\lambda_{\text{latent}} = 10^{-5}$ and $\lambda_{\text{aux}} = 10^{-4}$, for those  that use the respective loss components, and the learning rate is $10^{-4}$.
To evaluate the generated motion sequences against those in the dataset, we use the evaluation system (incl. pretrained classifier model for HumanAct12) established in and provided by \cite{chuan2020action2motion} as well as the extensions provided by \cite{petrovich21actor}. For the exact details of this evaluation system and classifier, we refer the reader to the two aforementioned works. 

\subsection{CL2Gen on HumanAct12 motion sequences}\label{sec_exp_incr_learn_humanact}
In this CL2Gen study of class-incremental conditional generation, the objective of the model is to learn to generate one or multiple classes in a given task and then move on to train on the next task. 
After each task and after training on all tasks, samples generated by the model are evaluated on metrics to assess the generative accuracy performance. For our experiments, we split the 12 action categories into 6 different tasks of 2 classes each. The model is first trained on a given task of 2 actions for 5000 epochs (5000 was chosen to be in line with the training procedures in \cite{chuan2020action2motion,petrovich21actor}). It is then used to generate samples of all classes encountered so far. These generated samples are combined with the training samples of the new task when training the model on the new task. We do not store generated samples used in previous tasks, thus, before training on a new task, new samples are generated for all previously seen classes. Therefore, given, \eg,  a class $a_1$ of task $k_1$, the generated replay samples used in $k_2$ are different than the ones used while training on task $k_3$ (see \ref{sec:mix_replay}). 
Following the observations in \cite{petrovich21actor}, the training samples have variable sequence durations between 60 and 100 frames. Testing is done on generated sequences with fixed length of 60 frames. 

For replay, we experiment with different numbers of replay samples that are of fixed length (60 frames).
We introduce a replay hyperparameter which defines the replay class ratio, \ie,

the number of replay samples generated per previously seen class compared to the number of new samples of the current task on which the network is being trained. Given a fixed number of training epochs per task, the more replay samples are added, the longer the training takes. 

The classes used in the individual tasks follow the default class list of HumanAct12. Task 1 consists of classes 'warm-up' and 'walk', task 2 of 'run and 'jump', and so on.

Experimental settings for results in Table \ref{humanact12_results} are as follows:
\begin{itemize}    
    \item replay class ratio of either $\nicefrac{1}{16}$ or $\nicefrac{1}{5}$, real samples with a random length between 60 and 100 frames and replay samples with a fixed length of 60 frames
    \item replay class ratio of either $\nicefrac{1}{16}$ or $\nicefrac{1}{5}$, real samples with a random length between 60 and 100 frames and replay samples with a fixed length of 60 frames, with auxiliary loss from a frozen pretrained classifier
    \item no replay samples. 
\end{itemize}

In some preliminary experiments, that are not part of this work, we tested with $\nicefrac{1}{64}$ replay as was used for non-sequential pure classification task in \cite{van2020brain}. Improvement was seen starting with a replay of $\nicefrac{1}{16}$, hence we chose this value. Additionally, to experiment for cases with even more replay samples, we arbitrarily chose $\nicefrac{1}{5}$. 

All experiments use AdamW~\cite{loshchilov2017decoupled}, 8-layer (256 units per layer) unidirectional GRUs for the en- and decoder, and $\bm{z} \in \R^{256}$.

\subsection{Offline vs. incremental learning results}
The results in Table \ref{humanact12_results} show that BI-CTVAE performs better in all metrics except FID when compared to the  offline GRU. Ground Truth are the metric values calculated on the actual dataset as reported by \cite{chuan2020action2motion}. With the trainable \gls{gmm} components BI-CTVAE has more available parameters and, therefore, an advantage for structuring the latent representation space. 
In the CL2Gen setting without auxiliary loss, the best model trained using replay samples shows an accuracy that is up to 3x higher than the baseline model without replay samples. Using real samples for replay improves FID and Diversity over models trained with generated replay samples. However, for accuracy and multimodality, using more generated replay samples lead to better results in these two metrics. The latter is preferred if the goal is to generate recognizable motions.

Similar to the BI-CTVAE results, the BI-CTVAE aux models show highest accuracy when using generated replay samples. In the case of real replay samples, BI-CTVAE models  perform better across all metrics, however in all other cases with generated replay samples, the accuracy and multimodality results from BI-CTVAE aux models are higher than those from BI-CTVAE. These results are also better than those of BI-CTVAE aux models that use real replay samples.

\begin{figure}
\begin{center}
\includegraphics[width=0.8\linewidth]{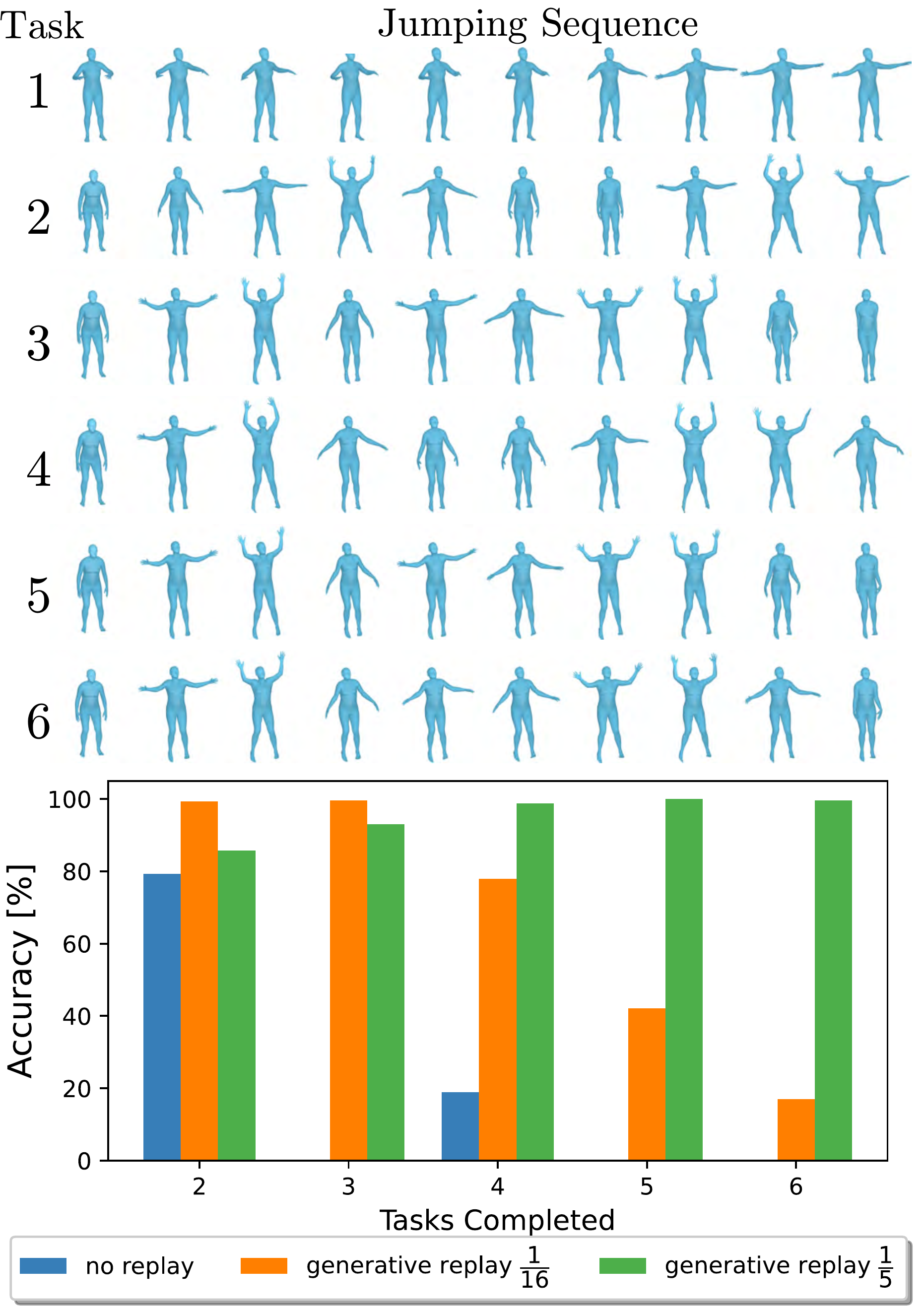}
\end{center}
\caption{\textbf{Top:} Generated `jump' action samples of the model trained with replay $\frac{1}{5}$ and auxiliary loss. After training on task 1 (`walk' + `warm up'), the model generates a motion similar to some 'warm up' samples. After training on `jump' in task 2, the model can generate this action. Even after training on the next tasks 3 to 6 with new action classes, the model retains its ability to generate realistic sequences for 'jump'. \textbf{Bottom:} classification accuracy of generated 'jump' samples after each task. Only the $\frac{1}{5}$ model maintains performance after training on subsequent tasks.}
\label{fig_task_samples}

\end{figure}

 Fig.~\ref{fig_task_samples} shows the samples generated for the "jump" class from a model trained incrementally starting from task 1. The first row shows that the model, if given the conditional input to generate a sample before learning to generate this class, will generate a sample which is similar to an already observed class. After learning the "jump" class in task 2, and by using replay samples during training of subsequent tasks, the model is able to still generate realistic sequences of the 'jump' class. 

\section{Conclusion}
This work presents a biologically-inspired generative replay model, BI-CTVAE, which performs well on our newly proposed continual learning to generate (CL2GEN) scenario for human motion.
The final classification accuracy of BI-CTVAE on the HumanAct12 dataset after sequentially learning all action classes is 78\%, which is 63\% higher than using no-replay, and only 5.4\% lower than a state-of-the-art offline trained GRU model.
Future work includes the testing of this model on other human motion datasets, so we can investigate  the use of this brain-inspired generative replay of sequences VAE model  in other CL2GEN scenarios. 

\vfill\pagebreak

\bibliographystyle{IEEEbib}
\bibliography{biovae_wacv_2022}

\begin{thebibliography}{10}

\bibitem{bremner2012multisensory}
Bremner, A.~J., Lewkowicz, D.~J., and Spence, C.,
\newblock {\em Multisensory development},
\newblock Oxford University Press, 2012.

\bibitem{tani2016exploring}
Tani, J.,
\newblock {\em Exploring robotic minds: actions, symbols, and consciousness as
  self-organizing dynamic phenomena},
\newblock Oxford University Press, 2016.

\bibitem{parisi2019continual}
Parisi, G.~I., Kemker, R., Part, J.~L., Kanan, C., and Wermter, S.,
\newblock ``Continual lifelong learning with neural networks: A review,''
\newblock {\em Neural Networks}, vol. 113, pp. 54--71, 2019.

\bibitem{mizuguchi2017changes}
Mizuguchi, N. and Kanosue, K.,
\newblock ``Changes in brain activity during action observation and motor
  imagery: their relationship with motor learning,''
\newblock {\em Progress in brain research}, vol. 234, pp. 189--204, 2017.

\bibitem{nakano2017understanding}
Nakano, H. and Kodama, T.,
\newblock ``Understanding neural mechanisms of action observation for improving
  human motor skill acquisition,''
\newblock in {\em Electroencephalography}. IntechOpen, 2017.

\bibitem{neuman2013direct}
Neuman, B. and Gray, R.,
\newblock ``A direct comparison of the effects of imagery and action
  observation on hitting performance,''
\newblock {\em Movement \& Sport Sciences-Science \& Motricit{\'e}}, , no. 79,
  pp. 11--21, 2013.

\bibitem{ste2012observation}
Ste-Marie, D.~M., Law, B., Rymal, A.~M., Jenny, O., Hall, C., and McCullagh,
  P.,
\newblock ``Observation interventions for motor skill learning and performance:
  an applied model for the use of observation,''
\newblock {\em International Review of Sport and Exercise Psychology}, vol. 5,
  no. 2, pp. 145--176, 2012.

\bibitem{ramanathan2015sleep}
Ramanathan, D.~S., Gulati, T., and Ganguly, K.,
\newblock ``Sleep-dependent reactivation of ensembles in motor cortex promotes
  skill consolidation,''
\newblock {\em PLoS biology}, vol. 13, no. 9, pp. e1002263, 2015.

\bibitem{cousins2016cued}
Cousins, J.~N., El-Deredy, W., Parkes, L.~M., Hennies, N., and Lewis, P.~A.,
\newblock ``Cued reactivation of motor learning during sleep leads to overnight
  changes in functional brain activity and connectivity,''
\newblock {\em PLoS biology}, vol. 14, no. 5, pp. e1002451, 2016.

\bibitem{mccloskey1989catastrophic}
McCloskey, M. and Cohen, N.~J.,
\newblock ``Catastrophic interference in connectionist networks: The sequential
  learning problem,''
\newblock in {\em Psychology of learning and motivation}, vol.~24, pp.
  109--165. Elsevier, 1989.

\bibitem{french1999catastrophic}
French, R.~M.,
\newblock ``Catastrophic forgetting in connectionist networks,''
\newblock {\em Trends in cognitive sciences}, vol. 3, no. 4, pp. 128--135,
  1999.

\bibitem{van2020brain}
van~de Ven, G.~M., Siegelmann, H.~T., and Tolias, A.~S.,
\newblock ``Brain-inspired replay for continual learning with artificial neural
  networks,''
\newblock {\em Nature communications}, vol. 11, no. 1, pp. 1--14, 2020.

\bibitem{cossu2021continual}
Cossu, A., Carta, A., Lomonaco, V., and Bacciu, D.,
\newblock ``Continual learning for recurrent neural networks: an empirical
  evaluation,''
\newblock {\em Neural Networks}, vol. 143, pp. 607--627, 2021.

\bibitem{ehret2020continual}
Ehret, B., Henning, C., Cervera, M.~R., Meulemans, A., von Oswald, J., and
  Grewe, B.~F.,
\newblock ``Continual learning in recurrent neural networks with
  hypernetworks,''
\newblock {\em arXiv preprint arXiv:2006.12109}, 2020.

\bibitem{sodhani2020toward}
Sodhani, S., Chandar, S., and Bengio, Y.,
\newblock ``Toward training recurrent neural networks for lifelong learning,''
\newblock {\em Neural computation}, vol. 32, no. 1, pp. 1--35, 2020.

\bibitem{ororbia2020continual}
Ororbia, A., Mali, A., Giles, C.~L., and Kifer, D.,
\newblock ``Continual learning of recurrent neural networks by locally aligning
  distributed representations,''
\newblock {\em IEEE transactions on neural networks and learning systems}, vol.
  31, no. 10, pp. 4267--4278, 2020.

\bibitem{sun2019lamol}
Sun, F.-K., Ho, C.-H., and Lee, H.-Y.,
\newblock ``Lamol: Language modeling for lifelong language learning,''
\newblock {\em arXiv preprint arXiv:1909.03329}, 2019.

\bibitem{duncker2020organizing}
Duncker, L., Driscoll, L., Shenoy, K.~V., Sahani, M., and Sussillo, D.,
\newblock ``Organizing recurrent network dynamics by task-computation to enable
  continual learning,''
\newblock {\em Advances in Neural Information Processing Systems}, vol. 33,
  2020.

\bibitem{kao2021natural}
Kao, T.-C., Jensen, K.~T., Bernacchia, A., and Hennequin, G.,
\newblock ``Natural continual learning: success is a journey, not (just) a
  destination,''
\newblock {\em arXiv preprint arXiv:2106.08085}, 2021.

\bibitem{tevet2022human}
Tevet, G., Raab, S., Gordon, B., Shafir, Y., Bermano, A.~H., and Cohen-Or, D.,
\newblock ``Human motion diffusion model,''
\newblock {\em arXiv preprint arXiv:2209.14916}, 2022.

\bibitem{cervantes2022implicit}
Cervantes, P., Sekikawa, Y., Sato, I., and Shinoda, K.,
\newblock ``Implicit neural representations for variable length human motion
  generation,''
\newblock {\em arXiv preprint arXiv:2203.13694}, 2022.

\bibitem{petrovich21actor}
Petrovich, M., Black, M.~J., and Varol, G.,
\newblock ``Action-conditioned 3{D} human motion synthesis with transformer
  {VAE},''
\newblock in {\em International Conference on Computer Vision (ICCV)}, 2021.

\bibitem{chuan2020action2motion}
Guo, C., Zuo, X., Wang, S., Zou, S., Sun, Q., Deng, A., Gong, M., and Cheng,
  L.,
\newblock ``Action2motion: Conditioned generation of 3d human motions,''
\newblock in {\em Proceedings of the 28th ACM International Conference on
  Multimedia (MM '20)}, 2020.

\bibitem{cho2014properties}
Cho, K., Van~Merri{\"e}nboer, B., Bahdanau, D., and Bengio, Y.,
\newblock ``On the properties of neural machine translation: Encoder-decoder
  approaches,''
\newblock {\em arXiv preprint arXiv:1409.1259}, 2014.

\bibitem{zhou2019continuity}
Zhou, Y., Barnes, C., Lu, J., Yang, J., and Li, H.,
\newblock ``On the continuity of rotation representations in neural networks,''
\newblock in {\em Proceedings of the IEEE/CVF Conference on Computer Vision and
  Pattern Recognition}, 2019, pp. 5745--5753.

\bibitem{loper2015smpl}
Loper, M., Mahmood, N., Romero, J., Pons-Moll, G., and Black, M.~J.,
\newblock ``Smpl: A skinned multi-person linear model,''
\newblock {\em ACM transactions on graphics (TOG)}, vol. 34, no. 6, pp. 1--16,
  2015.

\bibitem{higgins2016beta}
Higgins, I., Matthey, L., Pal, A., Burgess, C., Glorot, X., Botvinick, M.,
  Mohamed, S., and Lerchner, A.,
\newblock ``beta-vae: Learning basic visual concepts with a constrained
  variational framework,''
\newblock 2016.

\bibitem{zou20203d}
Zou, S., Zuo, X., Qian, Y., Wang, S., Xu, C., Gong, M., and Cheng, L.,
\newblock ``3d human shape reconstruction from a polarization image,''
\newblock in {\em European Conference on Computer Vision}. Springer, 2020, pp.
  351--368.

\bibitem{yang2019task}
Yang, G.~R., Joglekar, M.~R., Song, H.~F., Newsome, W.~T., and Wang, X.-J.,
\newblock ``Task representations in neural networks trained to perform many
  cognitive tasks,''
\newblock {\em Nature neuroscience}, vol. 22, no. 2, pp. 297--306, 2019.

\bibitem{loshchilov2017decoupled}
Loshchilov, I. and Hutter, F.,
\newblock ``Decoupled weight decay regularization,''
\newblock {\em arXiv preprint arXiv:1711.05101}, 2017.

\end{thebibliography}

\end{document}